%% file: Rank_Arxiv.tex
\documentclass[a4paper, 10pt]{article}
\usepackage{graphicx}

\usepackage{amsmath}
\usepackage{amsfonts}
\usepackage{amssymb}

\usepackage{xcolor}
\usepackage{hyperref}
\hypersetup{
    colorlinks,
    linkcolor={red!60!black},
    citecolor={blue!75!black},
    urlcolor={blue!80!red}
}
\usepackage{enumerate}
\usepackage{url}
\usepackage{enumitem}
\usepackage{graphicx}
\usepackage[numbers]{natbib} 
\usepackage{epsfig}
\usepackage[lofdepth,lotdepth]{subfig}

\usepackage{color}

\usepackage{algorithm}
\usepackage{algpseudocode}

\usepackage{verbatim}

\usepackage{bm}

\newcommand{\specialcell}[2][c]{%
  \begin{tabular}[#1]{@{}c@{}}#2\end{tabular}}
\newcommand{\bphi}{{\boldsymbol \phi}}
\newcommand{\bbeta}{{\boldsymbol \beta}}
\newcommand{\omegab}{{\boldsymbol \omega}}
\newcommand{\rhob}{{\boldsymbol \rho}}
\newcommand{\gpi}{(\nabla {\phi})^{-1}}

\newcommand{\bbR}{\mathbb{R}}
\newcommand{\bR}{\mathbf{R}}
\newcommand{\br}{\mathbf{r}}
\newcommand{\hr}{\bar{\mathbf{r}}}

\newcommand{\bz}{\mathbf{z}}
\newcommand{\hz}{\bar{\mathbf{z}}}

\newcommand{\bV}{\mathbf{V}}

\newcommand{\cV}{\mathcal{V}}

\newcommand{\bq}{\mathbf{q}}
\newcommand{\cQ}{\mathcal{Q}}

\newcommand{\bx}{\mathbf{x}}
\newcommand{\bX}{\mathbf{X}}
\newcommand{\by}{\mathbf{y}}

\newcommand{\dppi}[2]{D_{\bphi}\left({#1}\| \gpi({#2})\right)}

\newcommand{\dpi}[2]{D_{\bphi}\hspace*{-0.07cm}\left({#1}\| {#2}\right)}
\newcommand{\dpri}[2]{D_{\bphi_r}\hspace*{-0.07cm}\left({#1}\| {#2}\right)}
\newcommand{\dpzi}[2]{D_{\bphi_z}\hspace*{-0.07cm}\left({#1}\| {#2}\right)}

\setlist[itemize]{leftmargin=0.8cm}
\setlist[enumerate]{leftmargin=0.8cm}

\let\oldref\ref
\renewcommand{\ref}[1]{(\oldref{#1})}

\usepackage[margin=1.0in]{geometry}
\input econfmacros.tex
\begin{document}

\title{Monotone Retargeting for Unsupervised Rank Aggregation with Object Features}

\author{Avradeep Bhowmik\\[3pt] University of Texas at Austin\\ Austin, TX  \and Joydeep Ghosh\\[3pt] University of Texas at Austin\\ Austin, TX}

%\begin{titlepage}
%\pubblock
%
%\vfill
%\Title{Monotone Retargeting for Unsupervised Rank Aggregation with Object Features}
%\vfill
%\Author{ Despina Reggiano\support}
%\Address{\napoli}
%\vfill
%\begin{Abstract}
%I describe a case study of Mesmeric influence on a physiological  reaction
%in two  Albanian subjects.
%\end{Abstract}
%\end{titlepage}
%\def\thefootnote{\fnsymbol{footnote}}
%\setcounter{footnote}{0}
%%

\maketitle

\begin{abstract}
%intro
Learning the true ordering between objects by aggregating a set of expert opinion rank order lists is an important and ubiquitous problem in many applications ranging from social choice theory to natural language processing and search aggregation. We study the problem of unsupervised rank aggregation where no ground truth ordering information in available, neither about the true preference ordering between any set of objects nor about the quality of individual rank lists. Aggregating the often inconsistent and poor quality rank lists in such an unsupervised manner is a highly challenging problem, and standard consensus-based methods are often ill-defined, and difficult to solve. 
%body
In this manuscript we propose a novel framework to bypass these issues by using object attributes to augment the standard rank aggregation framework. We design algorithms that learn joint models on both rank lists and object features to obtain an aggregated rank ordering that is more accurate and robust, and also helps weed out rank lists of dubious validity. 
%expt
We validate our techniques on synthetic datasets where our algorithm is able to estimate the true rank ordering even when the rank lists are corrupted. Experiments on three real datasets, MQ2008, MQ2008 and OHSUMED, show that using object features can result in significant improvement in performance over existing rank aggregation methods that do not use object information. Furthermore, when at least some of the rank lists are of high quality, our methods are able to effectively exploit their high expertise to output an aggregated rank ordering of great accuracy.
\end{abstract}

\section{Introduction}\label{sec:intro}

Learning preference orderings among objects is a common problem in many modern applications including web search, document retrieval, collaborative filtering and recommendation systems. Rank aggregation is a version of this problem that appears in areas ranging from voting and social choice theory \cite{borda2}, to meta search and search aggregation \cite{meta} to ensemble methods for combining classifiers \cite{ensemble}. \\

We focus on the problem of unsupervised rank aggregation \cite{unsup1, unsup2} in this manuscript. The standard setup consists of a set of $n$ items or objects to be ranked, and a set of $p$ ranking lists from ``experts"\footnote{Can be IR features or learning based algorithms or subjective crowd-sourced annotation, for example} (possibly inconsistent, and of varying levels of expertise and credibility) containing rank scores or relevance scores for all or a subset of the $n$ items, and the objective is to combine the lists from different experts and obtain a consensus rank order over the set of items. Note that this setup is different from supervised rank aggregation \cite{supagg, sup2, sup3} or semi-supervised rank aggregation \cite{semi, semi2} where, in addition to the rank lists, we also have ground truth rank orderings between at least a subset of the objects to be ranked (for example, ground truth rank orderings supplied by high expertise human annotators or subject matter experts, that can then be used as training data). In contrast, in the unsupervised setup, we are only given access to rank lists without any information about the quality of each list, or any ground truth rank order between objects.\\

A related but very distinct problem that also looks at ordering among items is learning to rank or LETOR \cite{ndcg,rank1,rank2}, which tries to learn ranking functions over objects given training data with known rank scores or pairwise preference information. Standard methods for LETOR model the rank scores or orderings as a function of features associated with the objects. \\

The difference between LETOR and unsupervised rank aggregation is two-fold. First, unlike the latter, LETOR is supervised and has access to training data. %, while apart from certain restricted settings \cite{supagg,semi}, rank aggregation is an unsupervised learning problem which does not have access to the ``true" rank order. % {\color{red} beyond rank lists which are often of dubious credibility, and are almost never concurring}. 
Second, while LETOR models rank scores as functions of object features, existing rank aggregation methods are completely agnostic to the properties of the objects being ranked, even when such information is available, for example in the case of meta-search, search aggregation, or ensemble methods over learning agents\cite{meta,ensemble}. As a result, rank aggregation is significantly more challenging to tackle as compared to LETOR, or supervised/semi-supervised methods in general.\\

%The objective of this manuscript is to design a rank aggregation framework that uses the best of both worlds. Unlike LETOR, the framework is completely unsupervised and information about ``true" rank scores or pairwise preference information is not required. Furthermore, the framework augments standard rank aggregation by using information from object features.

Without access to any information about true preference ordering among objects, the standard rank aggregation problem becomes a hard combinatorial problem, and often rife with paradoxes \cite{arrow}. %Because of its unsupervised nature, this approach is rife \cite{paradox} with paradoxes, like Arrow's Impossibility Theorem \cite{arrow} which states that for every rank aggregation scheme over a set of more than two items at least one of the following three very natural and reasonable conditions are violated-Pareto efficiency among ranked lists, non-dictatorship, and independence of irrelevant items.% Pareto efficiency among ranked lists (if every expert prefers item $a$ over item $b$, the consensus should do the same), non-dictatorship (no individual expert should always determine the final ordering), and independence of irrelevant items (pairwise preference between two items should not be affected by the presence of a third item). 
Rule based approaches like Borda winner \cite{borda}, Condorcet winner \cite{condorcet}, etc., are commonly used, but many of these rules are incompatible with each other \cite{borda2}. Alternative approaches learn consensus orderings by minimising ``disagreements" or distance-like metrics among the rank scores or preference orderings specified by the ranked lists from experts \cite{kdt, spr}. This approach runs into the problem that for many of the commonly used distance metrics, the corresponding optimisation problem is NP-Hard \cite{dwork}.\\

Indeed, without additional information, one cannot form an incontrovertible definition for ``consensus" and ``disagreement" between rank scores, or even decide on the metric to be used to find the consensus rank ordering. Moreover, these methods are excessively dependent on the assumption that the experts are reliable and competent. Spurious rank order lists or even the presence of noise in the ranked lists can significantly deteriorate the performance of most standard methods used for rank aggregation. 

\subsection{Using Object Information}

%The preceding discussion makes it clear that, in contrast to LETOR, 
The absence of supervision reduces rank aggregation to a problem of choosing between competing heuristics. But obtaining even partial ground truth can be expensive and time consuming and often requires the involvement of dedicated human experts as annotators.  \\
 
This manuscript introduces a novel rank aggregation framework that mitigates the handicaps of unsupervised rank aggregation by using information about object attributes to augment standard approaches and aid the recovery of the ``true" rank order. This is in contrast to existing methods 
which are ``blind" in the sense that they are completely agnostic to the properties of the objects themselves.\\

Our key contributions are summarised as follows-\vspace*{-0.1cm}
\begin{enumerate}
\item To the best of our knowledge, we are the first to use object information to aid the rank aggregation problem. We introduce a novel framework that combines information from rank scores and object features to learn a consensus ordering over a set of items.\vspace*{-0.2cm}
%first to do rank aggregation with object features
\item We formulate the rank aggregation problem using isotonically coupled models over expert lists and object features to model rank scores, and describe a solution algorithm to estimate the true ordering that involves alternate but interdependent iterations between a LETOR step and a rank aggregation step, each of which is separately a simple convex optimisation problem with monotonicity constraints.\vspace*{-0.2cm}
\item We evaluate our framework with experiments on synthetic data where, unlike existing methods, our algorithm is able to reconstruct the true rank ordering exactly given corrupted rank lists. We also demonstrate our methods on three real datasets, where our algorithm significantly outperforms existing rank aggregation techniques that do not use object information.\vspace*{-0.1cm}
\end{enumerate}

%At this point it might be useful to distinguish between ranking and rank aggregation. While supervised variations for rank aggregation exist (for instance, \cite{comb2}), for most applications, rank aggregation is completely unsupervised in the sense that the true rank scores are not known. On the other hand, ranking or learning to rank (LETOR) is a supervised method that, given training data involving object features and a single list of rank or relevance score, learns a ranking function that returns a relevance score from the object features (see for instance, \cite{ndcg}, \cite{rank1}, \cite{rank2}, \cite{rank3} and references therein). Supervised rank aggregation \cite{supagg} or semi-supervised ranking methods \cite{semi} also use partial or full supervision in the form of scores or pairwise preference information for items.

We note that even though our framework uses LETOR inspired methods, our formulation is still an unsupervised learning algorithm since we do not assume access to any full or partial ground truth rank order or pairwise preferences, or even information about the quality of any rank lists.

%Arrow's Theorem
%Metrics
%NP-hard
%\subsection{Related Work}

%\begin{comment}
\subsection{Related Work}\label{sec:related}

%The methods outlined in this manuscript involve techniques from monotone retargeting \cite{MR, memr} that involves learning ranking functions using monotonically transformed linear functions of object features. A general discussion on monotone retargeting is deferred to Appendix \ref{MR}. Here, we describe standard methods for rank aggregation commonly used in most real world applications.

%Alternative criteria, like Borda winner \cite{borda}, Condorcet winner \cite{condorcet}, etc. have been used, many of which are incompatible with each other \cite{borda2} leading to further problems. %In many cases, a winner may not even exist.

%In light of fundamental contradictions in defining rules for rank aggregation, alternative approaches use measures of similarity (or alternatively, distance-like metrics) between orderings or permutations \cite{kdt}\cite{lbd}\cite{spr}. The consensus ordering is then obtained by solving an optimisation problem to find an ordering that maximises the similarity measures between the consensus ordering and each of the orderings specified by the ranked lists from experts. However, for many of the commonly used distance metrics the optimisation problem is NP-Hard \cite{dwork}. 

In light of the difficulties inherent in the rank aggregation problem, most of the commonly used rank aggregation applications apply heuristic based techniques ranging from classical methods for vote counting, combination methods involving linear and non-linear functions, and more recent approaches involving Markov Chain Monte Carlo simulations.\\% Methods have been developed which can work on either rank scores or positional\\

The Borda Count \cite{borda} is the traditional vote counting procedure that uses positional information as opposed to rank scores. %Versions of Borda Count have been used throughout history and are still being used at various levels of administrative elections. %The basic idea of the Borda Count is to get preferential orderings over the entire set of candidates from every voter. For each voter, each candidate is then assigned positional scores according to the number of candidates ranked below them. The candidate which has the highest sum of positional scores among all voters is then declared the winner. 
The procedure orders the candidates (items) by the number of candidates ranked lower than them, averaged over the set of voters (expert lists).\\ %Interestingly, Borda Count is the only voting procedure that satisfies all of the symmetry properties that should be satisfied by a reasonable voting procedure \cite{borda2}.

Combination methods \cite{comb1,comb2} work on explicit relevance scores provided by experts and include various linear combinations like CombSUM, CombMNZ and CombANZ as well as non-linear methods like CombMIN and CombMAX. 
%Linear combination methods estimate the rank order as defined by a simple sum of rank lists by experts (CombSUM), or a weighted sum using different weighting strategies (CombMNZ and CombANZ) depending on the number of rank lists that retrieve each item.
%In CombSUM, for each item the relevance scores from each expert is added and the items are ranked according to the sum of their scores. CombMNZ is a variation in which the sum of scores are further weighted by the number of expert lists that retrieve the particular item, and the ranking is determined from the weighted sum of scores. CombANZ is exactly the same except the weights used are the reciprocal of the number of rank lists that retrieve the item. 
It has been seen \cite{comb2,comb3} that out of the three linear combination methods, CombMNZ has the best performance. If all items have been retrieved by all expert lists, all three methods give the same final ranking.\\

Among non-linear methods \cite{comb2,comb1}, CombMIN and CombMAX respectively rank the items on the minimum and maximum scores received across the lists provided by experts.\\

More recent work \cite{dwork} on rank aggregation have introduced MCMC based methods. The basic idea is to use the items to be ranked as states of a Markov Chain and define the transition probability of switching from one state (item) to another, based on the relative scores or preference values of the corresponding items across rank lists. Four different Markov Chain constructions (MC1, MC2, MC3, MC4) have been described in \cite{dwork} that use different heuristics to construct the transition probability matrix. The final ranking of the items is defined by the stationary distribution across the items defined by the states of the Markov Chain.\\ %That is, items which have a higher weight in the stationary distribution are ranked higher.% and vice versa.% The basic idea is that items to be ranked higher get a higher weight.
%\end{comment}
\vspace*{0.1cm}
\noindent\textbf{Note On Notation:} The vector $\mathbf{r}$ is said to be in increasing order if $r_{i} \leq r_{j}$ whenever $i \leq j$. The set of all such vectors in $\mathbb{R}^n$ is denoted with a subscripted downward pointing arrow as $\mathbb{R}^n_\downarrow$. Two vectors $\mathbf{r}$ and $\mathbf{z}$ are said to be isotonic, $\mathbf{r \sim_\downarrow z}$, if $r_{i} \geq r_{j}$ if and only if $z_{i} \geq z_{j}$ for all $i,j$.%\vspace*{-0.1cm}

\section{Problem Setup}

Let $\cV$ be a set of distinct items and $\cQ = \{\bq_1, \cdots , \bq_{|\cQ|}\}$ be a set of queries. Each query $\bq$ is associated with a set of items $\bV_\bq \subset \cV$. %For simplicity, assume that $|\bV_\bq| = n$ for each query $\bq$. 
For simplicity and with no loss in generality, we consider the case of a single query over a set of $n$ items, and drop the notation $\bq$ from all subsequent notation.\\

Suppose the true rank ordering over $n$ items is given by a vector $\rhob^* \in \bbR^n$. In search aggregation, for example, $\rhob^*$ could be true relevance scores as annotated by a human subject matter expert\footnote{we study rank aggregation in the context of applications like meta-search and IR where such a $\rhob^*$ is assumed to exist, as opposed to social choice theory where such a $\rhob^*$ may not always exist because of Arrow's Impossibility Theorem \cite{arrow}}. With slight abuse of notation, we shall overload $\rhob^*$ to denote both a rank score as well as a rank ordering (permutation) defined by the rank score vector.\\

In the standard rank aggregation setup, we are given access to rank lists by a set of $p$ experts, each of whom assert a rank ordering or relevance score judgement over the entire set of items in $\bV$. Suppose that the relevance score vector for the $k^{th}$ expert is $\br^k \in \bbR^n$, where a higher score indicates a higher relevance for the item. We aggregate all relevance score lists as columns of the rank list matrix $\bR = [\br^1; \br^2; \cdots \br^p] \in \bbR^{n \times p}$.\\

In unsupervised rank aggregation, the true ordering $\rhob^*$ is not known, even among a partial subset of objects. However, it is assumed that at least some of the rank lists in $\bR$ are generated using $\rhob^*$ (perhaps from noisy or corrupted versions of $\rhob^*$) and standard rank aggregation methods try to recover $\rhob^*$ by learning a model that maps $\bR$ to a vector $\hr$ that is isotonic with $\rhob^*$ (e.g. Borda Count, combination methods, etc. see section \ref{sec:related}).\\% The usual strategy is to use consensus based heuristics to find an $\hr$ that minimises disagreements. However, without additional information, it is difficult to even form an incontrovertible definition for ``disagreement" between rank scores. Additionally these approaches is susceptible to degradation in performance in case of spurious rank orders or excessive noise in $\bR$.

In this manuscript we try to overcome the limitations of unsupervised rank aggregation as described in section \ref{sec:intro} by augmenting our setup with object features, which are often available in applications like meta-search, IR, etc. Suppose for the set of items $\cV = \{V_1, V_2, \cdots V_n\}$, each item $V_i$ can be represented by a $d$-dimensional feature vector $\bx_i \in \bbR^d$. We collect all these item feature vectors to form rows of the feature matrix $\bX \in \bbR^{n \times d}$. The consensus rank ordering is then obtained using information from both $\bR$ and $\bX$.\\

To motivate our approach, consider the standard supervised LETOR framework which has access to both $\bX$ as well as $\rhob^*$. Popular LETOR methods \cite{rank1, rank2, rank4} proceed by learning a model that maps $\bX$ to a rank score vector $\hz$ that is isotonic with $\rhob^*$. %Now consider the supervised rank aggregation framework where $\bR$ is known and full or partial information about $\rhob^*$ is available. 
Similarly, many standard methods for supervised rank aggregation \cite{supagg, sup2} proceed by learning a model that maps $\bR$ to a vector that agrees with available information about $\rhob^*$.\\

In contrast, $\rhob^*$ is completely unknown in unsupervised rank aggregation. The key idea for our method is that while we do not have explicit access to $\rhob^*$, both $\bX$ and $\bR$ are implicitly tied together by $\rhob^*$, and this implicit association can be exploited to recover $\rhob^*$ by jointly modelling the mappings that can be learned from $\bX$ and $\bR$ respectively.\\

To our knowledge, while object attributes are commonly available for many rank aggregation setups (e.g. meta-search or search aggregation), none of the existing rank aggregation methods exploit them sufficiently. As a first work, we explore the use of generalised linear models or GLMs \cite{glm1989} as our modeling framework since they are a large class of models that subsume many standard modeling frameworks (e.g. Gaussian or Poisson regression), and are widely used for many applications across a wide variety of domains. In particular, GLMs have found extensive usage in both LETOR \cite{MR, memr} as well as most\footnote{specifically, many common rank aggregation models, e.g. Borda, CombMNZ, etc. use simple linear schemata} rank aggregation methods \cite{borda, comb1, comb2}.

\section{Rank Aggregation with Object Features}

GLM's model the target variable (in this case, rank score) as a linear function of the feature variables, monotonically transformed via a monotonic link function. We assume that the true rank ordering ${\boldsymbol \rho}^*$ can be modeled as a monotonically transformed linear combination of the rank order lists $\bR \in \bbR^{n \times p}$ by the experts, that is, ${\boldsymbol \rho}^* \sim_\downarrow \bR\bbeta^*$ for some $\bbeta^* \in \bbR^p$. Even without the monotonic transformation, this setup subsumes standard score fusion algorithms, eg. see \cite{LBR, comb1}, and also the Borda-Count and weighted Borda-Fuse \cite{borda, borda2} if the rank scores are defined as the number of items ranked below a particular item. This setup can also effectively handle the case when some of the ranked lists are of dubious validity-- such rank lists can simply be assigned zero weight and discarded by the model.\\

Similarly, we also assume that the true rank score vector ${\boldsymbol \rho}^*$ is isotonic to a monotonically transformed linear function of the object features, that is, ${\boldsymbol \rho}^* \sim_\downarrow \bX\omegab^*$ for some $\omegab^* \in \bR^d$. This is a standard assumption for many ranking problems that model the rank function using a generalised linear model (see \cite{MR,memr} and references). \\

The objective now is to estimate $\bbeta^*$ and $\omegab^*$ by joint modelling of $\bX\omegab^*$ and $\bR\bbeta^*$ to estimate $\omega^*$ and $\bbeta^*$ by minimising an appropriate distance-like cost function. Clearly, standard cost functions like square loss may not make sense in every context (for example, when the domain is binary, or integer valued, or categorical), therefore we present our techniques for a much more general class of cost functions called Bregman divergences, which are distance-like functions intimately associated with GLM's and include many standard and commonly used loss functions like square loss, KL-Divergence, Generalised I-Divergence, etc.\\

\textbf{Bregman Divergences:} The matching loss functions associated with learning GLM parameters are distance-like functions called Bregman divergences, which are generalisations of square loss. Bregman Divergences are always defined on a convex function $\bphi(\cdot)$, where the particular convex function used depends on the particular GLM model used (see \cite{banerjee}). For any two vectors $\by$ and $\bx$, the Bregman divergence $D_\bphi(\cdot \| \cdot)$ between the vectors corresponding to the function $\bphi$ is defined as
\begin{equation*}
D_\bphi\left(\mathbf{y} \| \bx\right) \triangleq \bphi(\mathbf{y}) - \bphi(\bx) -  \langle \nabla\bphi(\bx), \mathbf{y} - \bx \rangle
\end{equation*}
A more detailed description of Bregman Divergences is deferred to Appendix \ref{sec:Bregman}, see also \cite{banerjee} for a rigorous exposition on the relationship between GLM's and Bregman Divergences. In particular, for our work we use the fact that parameter estimation in a GLM with given object features $\mathbf{X}$ and known target variable $\bz$ is equivalent to finding the minimiser ${\bm \omega}^*$ for $D_\bphi\left(\mathbf{z} \| \gpi(\mathbf{X}{\bm \omega})\right)$ as,

\begin{equation}\label{eq:stdglm}
\omegab^* = \text{arg}\min_{\omegab} \dppi{\bz}{\bX\omegab}
\end{equation}

where $\bphi(\cdot)$ is the convex function associated with the specific GLM used-- for a Gaussian model or equivalently square loss, for example, $\bphi$ is the identity function (see \cite{banerjee}).  \\

\noindent Bregman Divergences are always non-negative $D_\bphi(\cdot \| \cdot) \geq 0$, and by definition, $\dppi{\bz}{\bX\omegab}$ is separately convex in $\bz$ as well as in ${\bm \omega}$. Many standard distance-like functions like Square loss, Kullback-Leibler (KL) divergence and Generalised I-divergence can be written as members of this family for their corresponding $\bphi$. For succinctness, we shall henceforth denote $\dppi{\bz}{\bX\omegab}$ simply as $D_\bphi(\mathbf{z} \| \mathbf{X}{\bm \omega})$.

%Assume that there exists two vectors $\bbeta^*$ and $\omegab^*$ for the object attributes and the rank lists respectively, such that the true rank order is isotonic with $\bR\bbeta^*$ as well as $\bX\omegab^*$.

\subsection{Cost Function}

Following the discussion on using Bregman divergences as the loss function for estimating $\omegab^*$ and $\bbeta^*$, it is tempting to use $\bR\bbeta$ as the target variable and $\bX\omegab$ as the feature map, and write the joint optimisation framework as follows:

\begin{equation}\label{eq:joint1}
\bbeta^*, \omegab^* = \text{arg}\min_{\bbeta, \omegab} \dpi{\bR \bbeta}{\bX\omegab}
\end{equation}

%Alternatively, the order of arguments can be reversed to get the following optimisation problem
%\begin{equation}\label{eq:joint2}
%\bbeta^*, \omegab^* = \text{arg}\min_{\bbeta, \omegab} \dpi{\bX\omegab}{\bR \bbeta}
%\end{equation}

Alternatively, the order of arguments $(\bR \bbeta, \bX\omegab)$ can be reversed. However, both these formulations are deficient in their modelling capacity since they force a coupling between the domain of $\bX$ and the domain of $\bR$. In particular, while $\bX$ is often real valued, $\bR$ can often be integer valued or binary valued, or even categorical with partial ordering and the same $\bphi$ may not make sense for both. Moreover, this does not take into account the more general assumption that the true ordering need only be isotonic to the two linear functions, that is $\bR \bbeta \sim_\downarrow \bX\omegab$, but exact equality may be an unnecessarily strong and superfluous constraint. Incorporating monotonic invariance between $\bR\bbeta$ and $\bX\omegab$ is a challenging problem in the above formulation. \\

A far better alternative scheme that bypasses all these issues involves decoupling the cost function into two parts- one involving the rank scores by experts and one involving the object features.

\subsubsection{Decoupled Cost Function}

Consider $\hr \in \bbR^n$ as the rank score vector to be fitted against a linear function of the rank score matrix $\bR$. This part of the decoupled cost function therefore consists of the term $\dpri{\hr}{\bR\bbeta}$ for an appropriate $\bphi_r$. Since we consider all rank score vectors that invoke the same ordering among items to be equivalent, we shall be learning $\hr$ across all monotonic transformations to incorporate invariance across isotonic vectors in our formulation.\\ %Since this part involves coalescing the rank score from experts, we call this part of the cost function the rank aggregation cost function.

Similarly, suppose $\hz \in \bbR^n$ is the rank score vector obtained as the linear function of object features. Correspondingly, this part of the cost function becomes $\dpzi{\hz}{\bX\bbeta}$ for an appropriate $\bphi_z$. Along the same lines as above, we learn $\hz$ across all monotonic transformations.\\

Therefore, the full cost function becomes 
\begin{equation}\label{eq:overall}
\mathcal{C}(\br, \bbeta, \hz, \omegab) = \dpri{\br}{\bR\bbeta} + \lambda \dpzi{\hz}{\bX\omegab}
\end{equation}
%$$\mathcal{C}(\hr, \bbeta, \hz, \omegab) = \dpri{\hr}{\bR\bbeta} + \dpzi{\hz}{\bX\omegab}$$
to be minimised over $\hr, \bbeta, \hz, \omegab$. To retain the isotonicity relationship between the linear functions $\bR\bbeta$ and $\bX\omegab$, we add the constraint $\hr \sim_\downarrow \hz$. The overall optimisation problem therefore becomes
\begin{equation} \label{eq:full}
\begin{aligned}
\underset{\hr, \bbeta, \hz, \omegab}{\text{min}}\ \ & \ \ \dpri{\hr}{\bR\bbeta} + \lambda \dpzi{\hz}{\bX\omegab} \\
\text{s.t. }  & \ \ \hr \sim_\downarrow \hz
\end{aligned}
\end{equation}

\noindent \textbf{Choice of divergence} functions $\bphi_r$ and $\bphi_z$ depends on the domain (real valued, integer, etc.) and the modeling assumptions used on $\hr$ and $\hz$, a discussion on learning the appropriate $\bphi$ has been detailed in \cite{learnphi}. We shall see later that our optimisation framework involves steps that are invariant to the $\lambda$ value chosen, so we use $\lambda = 1$ for simplicity.

\section{Monotonically Retargeted Rank Aggregation}

%We assume that the true rank ordering ${\boldsymbol \rho}^*$ can be modeled as a monotonically transformed linear combination of the rank order lists $\bR \in \bbR^{n \times p}$ by the experts, that is, ${\boldsymbol \rho}^* \sim_\downarrow \bR\bbeta$ for some $\bbeta \in \bbR^p$. Even without the monotonic transformation, this setup subsumes standard score fusion algorithms, eg. see \cite{LBR} and \cite{comb1}, and also the Borda-Count and weighted Borda-Fuse if the rank scores are defined as the number of items ranked below a particular item. This setup also contains the case when some of the ranked lists are of dubious validity.

%Following the MR framework, given object features $\bX \in \bbR^{n \times d}$ we also assume that the true rank score vector ${\boldsymbol \rho}^*$ is isotonic to a monotonically transformed linear function of the object features, that is, ${\boldsymbol \rho}^* \sim_\downarrow \bX\omegab$ for some $\omegab \in \bR^d$. This is a standard assumption for many ranking problems that model the rank function using a generalised linear model (see \cite{MR,memr} and references)

Equation (\ref{eq:full}) is an optimisation problem of a function that is separately convex in its arguments, but over a non-convex set. Joint optimisation over all variables simultaneously is difficult, therefore we divide the variables into two disjoint sets and perform alternating minimisation over each set of variables separately. Nevertheless, because of the monotonic invariance of the isotonicity constraint we need to perform each optimisation step over all monotonic transformations of the vectors $\hr$ and $\hz$, which is a very difficult problem in general. However, it turns out that the setup becomes relatively easy to handle by using a technique called monotone retargeting used in supervised learning to rank problems.\\

\textbf{Monotone Retargeting} or \textbf{MR} \cite{MR} is a LETOR technique that uses object attributes $\bX$ and known rank score vector $\rhob^*$ and learns a model that finds the best mapping from $\bX$ over all possible monotonic transformations $\bz$ of the rank score vector $\rhob^*$. The specific optimisation problem that MR handles in the case of Bregman Divergences (i.e., using GLM's) is the following:

\begin{equation} \label{eq:mr}
\begin{aligned}
&\underset{\bz, \omegab}{\text{min}} & & D_\bphi\left(\mathbf{z} \| \gpi(\mathbf{X \omegab})\right) \\
& \text{s.t. } & & \bz \sim_\downarrow {\boldsymbol \rho}^*
\end{aligned}
\end{equation}

To impose strict ordering constraints and to avoid degenerate solutions, a variation of MR called Margin Equipped Monotone Retargeting or MEMR was introduced in \cite{memr} that uses margin constraints on $\bz$ together with ordering constraints defined by $\rhob^*$. A detailed description of MR and MEMR is given in Appendices \ref{sec:MR} and \ref{sec:memr} respectively. The steps used in the exact optimisation algorithm for MR is summarised as Algorithm \ref{alg:mr}.\\

MR is a versatile framework that has found usage outside of the traditional supervised ranking application it was designed for. In particular, versions of this framework have been used for collaborative filtering for recommendation systems \citep{MRMF}, and learning generalised linear models from aggregated data \cite{glmagg}. In this manuscript we apply this framework to the problem of rank aggregation.

\begin{algorithm}
\caption{LETOR with Monotone Retargeting}\label{alg:mr}
\begin{algorithmic}[1]
\Procedure{MR}{$\bphi, \bX, {\boldsymbol \rho}^*$}%\Comment{The g.c.d. of a and b}
\State Initialise $\omegab, \hz$
\While {not converged}
\State Solve for $\hz^+$ using PAV \vspace*{-0.1cm} $$\hz^+ = \text{arg}\min_{\hz \sim_\downarrow {\boldsymbol \rho}^*} \dpi{\hz}{\bX \omegab}\vspace*{-0.1cm}$$
\State Solve for $\omegab^+$ a std GLM param estimation\vspace*{-0.1cm}
$$\ \ \ \omegab^+ = \text{arg}\min_{\omegab} \dpi{\hz^+}{\bX \omegab}\vspace*{-0.15cm}$$
\State Update variables $(\hz, \omegab) = (\hz^+, \omegab^+)$
\EndWhile
%\State 
\State \textbf{return} $\hz, \omegab$%\Comment{Sparse map is }
\EndProcedure
\end{algorithmic} 
\end{algorithm}

Consider the set of variables $\Upsilon = (\hr, \bbeta, \hz, \omegab)$ divided into two sets of variables $\Upsilon_\br = (\hr, \bbeta)$ and 
$\Upsilon_{\bz} = (\hz, \omegab)$. After initialisation (which can be done by using any preferred rank aggregation algorithm), the algorithm proceeds in two alternating steps. First, keeping $\Upsilon_{\bz}$ fixed, the optimisation problem over $\Upsilon_\br$ becomes
\begin{equation} \label{eq:agg}
\begin{aligned}
&\underset{\hr, \bbeta}{\text{min}} & & \dpri{\hr}{\bR\bbeta} \\
& \text{s.t. } & & \hr \sim_\downarrow \hz
\end{aligned}
\end{equation}

Similarly, keeping $\Upsilon_{\br}$ fixed, the optimisation problem over $\Upsilon_\bz$ becomes
\begin{equation} \label{eq:rank}
\begin{aligned}
&\underset{\hz, \omegab}{\text{min}} & & \dpzi{\hz}{\bX\omegab} \\
& \text{s.t. } & & \hz \sim_\downarrow \hr
\end{aligned}
\end{equation}

At face value, because of the common isotonicity constraint $\hr \sim_\downarrow \hz$ applied to both steps of the optimisation problem, it may seem the algorithm will remain perpetually stuck to the rank ordering defined by the initialisation of $\hr$ or $\hz$. However, this does not happen because at alternate steps, the $\hz$ or $\hr$ that define the isotonicity constraint can be partially ordered rather than totally ordered. This enables the algorithm to freely move between permutations at successive steps. A detailed discussion of this phenomenon is provided in Section \ref{moveperm}.

\begin{algorithm}
\caption{Rank Aggregation with Object Features}\label{alg:mragg}
\begin{algorithmic}[1]
\Procedure{MR-RankAgg}{$\bphi_r, \bR, \bphi_z, \bX$}%\Comment{The g.c.d. of a and b}
\State Initialise $\hr, \bbeta, \hz, \omegab$
\While {not converged}\vspace*{0.1cm}
\State $\hz^+, \omegab^+ = \text{\textbf{MR}}(\bphi_z, \bX, \hr)$ \dotfill \textsc{LETOR-step}\label{z-step} \vspace*{0.1cm}
\State $\hr^+, \bbeta^+ = \text{\textbf{MR}}(\bphi_r, \bR, \hz^+)$ \dotfill \textsc{Rank-Agg step}\label{r-step} \vspace*{0.1cm}
\State Update all variables\vspace*{-0.1cm} $$ (\hr, \bbeta, \hz, \omegab) \leftarrow  (\hr^+, \bbeta^+, \hz^+, \omegab^+) \vspace*{-0.15cm}$$
%\begin{eqnarray*}
%\hr, \bbeta &\leftarrow & \hr^+, \bbeta^+ \vspace*{-0.1cm} \\
%\hz, \omegab &\leftarrow & \hz^+, \omegab^+ \vspace*{-3cm}
%\end{eqnarray*}
%$(\hr, \bbeta) = (\hr^+, \bbeta^+)$ and $(\hz, \omegab) = (\hz^+, \omegab^+)$
\EndWhile \vspace*{0.1cm}
%\State 
\State \textbf{return} $(\hr, \bbeta, \hz, \omegab)$%\Comment{Sparse map is }
\EndProcedure
\end{algorithmic}
\end{algorithm}

Individually, both equations (\ref{eq:agg}) and (\ref{eq:rank}) are standalone instances of MR and inherit all the desirable properties of the algorithm as detailed in \cite{MR, memr}. The two main steps of the algorithm have a nice interpretation-- equation (\ref{eq:agg}) is equivalent to MR($\bphi_r, \bR, \hz$) and is analogous to a LETOR step, while equation (\ref{eq:rank}) is equivalent to MR($\bphi_z, \bX, \hr$) and is analogous to a rank aggregation step. The steps used in the overall optimisation are summarised in Algorithm \ref{alg:mragg}. \vspace*{0.2cm}

%Therefore, the overall framework for rank aggregation with object features inherits all the desirable properties of LETOR with MR as detailed in \cite{MR} and \cite{memr}. 
%Note that equation \ref{eq:rank} is analogous to a LETOR problem for learning a rank function while \ref{eq:agg} is analogous to a rank aggregation problem. A summary of the steps is provided as Algorithm \ref{alg:mragg}.

\noindent\textbf{Convergence and Efficiency:} Algorithm \ref{alg:mragg} uses alternating minimisation on a non-negative cost function, therefore the algorithm always converges to a stationary point. Each iteration involves the MR procedure detailed in Algorithm \ref{alg:mr}, which can be implemented very efficiently in practice (see \cite{memr}). In particular, the PAV step has been implemented in $\Theta(n)$ (see references in \cite{iso}). Both the PAV step and the GLM-solver step have been extensively studied, and fast off-the-shelf implementations are readily available.%\vspace*{-0.1cm}

\subsection{Margin Equipped Version}\label{sec:margin}

%Similar to the margin equipped version of monotone retargeting (MEMR), a margin equipped version can be designed for this rank aggregation scheme as well. However, the margin constraints of MEMR cannot be used directly.

To avoid certain kinds of degeneracies (see \cite{memr}), MEMR uses a formulation that enforces an $\epsilon$-margin between the relevance scores between any two items adjacent to each other in the learned rank score vector $\bz$. We can formulate a margin-equipped version for our methods as well, however the MEMR scheme cannot be used directly. As we discuss in section \ref{moveperm}, one of the salient desiderata of our formulation is to be able to move between rank orderings at every step of the algorithm, which is achieved by using the partial orderings for $\hr$ or $\bz$ generated by the optimisation algorithm at the end of each iteration. Imposing a strict ordering would make that no longer possible.\\ %For instance, suppose we initialise $\hr$ and solve an MEMR optimisation problem to obtain $\hz^+$ with the rank ordering specified by $\hr$, enforcing a margin constraint on $\hz^+$ would force the total ordering defined by the initial $\hr$ on the final $\hz^+$ obtained. Subsequently, since $\hr^+$ would be optimised under a total order constraint specified by $\hz^+$, we end up with the same total ordering for both $\hr$ and $\hr^+$; thus, the algorithm will never be able to get out of the ordering scheme defined by the initialisation. 

%This can be avoided by simply enforcing a margin between the maximum and the minimum rank score instead of a margin between every adjacent rank score. 
In fact, in most cases a strict ordering is unnecessary-- to avoid degenerate solutions as in \cite{memr}, it is sufficient to enforce a margin only between the maximum and the minimum rank score. That is, the constraint we shall use, say on $\hr$, would be of the form $\hr^{(max)} - \hr^{(min)} > \epsilon$ for some $\epsilon > 0$. While this is a non-linear, non-convex condition in general, within the constraint set $\hr \sim_\downarrow \hz$, this becomes a simple half-space constraint since the \emph{indices} for maximum and minimum in $\hr$ are already specified by the ordering defined by $\hz$. A similar margin constraint can be applied to $\hz$ as well.\vspace*{-0.1cm}

\section{Discussion}

\subsection{Regularisation} %In the interest of brevity, so far we presented our cost function without an explicit regulariser on $\bbeta$ and $\omegab$. 
%It is trivial to see that adding any convex regularisation would not affect our overall methods in any way. 
Adding a convex regularisation term (especially on the GLM parameters $\bbeta$ and $\omegab$) in equation \ref{eq:full} can have many useful effects while retaining convexity properties in the optimisation problem. In particular, suppose we know that some of the rank order lists are spurious or generated by sources of dubious expertise. In such a case, sparsity promoting methods like LASSO regularisation on $\bbeta$ may help weed out the spurious rank lists by enforcing sparsity. A similar argument can also be made for adding sparsity boosting regularisers on $\omegab$ to weed out spurious features.\\ %In fact, with appropriate regularisation (see \cite{memr}) each portion of the cost function can be made jointly convex in all variables.

\subsection{Moving between permutations} \label{moveperm}
%Given a permutation, the set of all vectors in $\bbR^n$ ordered according to that permutation is a convex cone (see \cite{MR} for a more detailed exposition). 
One of the key desiderata of our algorithm is that the $\hr$ or $\hz$ in intermediate steps should be free to move between permutations, so that the final aggregated output for rank order does not get influenced too heavily by the initialisation. This is accomplished by our algorithm in the following manner. \\

The update step for $\hz$ in the MR Algorithm \ref{alg:mr} involves a pool-adjacent-violator (PAV) smoothing operation, which has the property that if the ordering constraint on $\hz$ does not match the ordering of the right hand side $\bX\bbeta$, the final output would be a partially ordered $\hz$. %Moreover, when such a partially ordered ${\boldsymbol \rho}$ is used as an input to MR, an intermediate step involves a permutation estimation which can impose a total ordering (consistent with the ordering on ${\boldsymbol \rho}$) on the final $\hz$ obtained. % If this partial ordering is then used as input to an MR step, an intermediate 
Suppose at time $t$, we start with some total ordering $\rhob_0$ on $\hz$ and $\hr$, and after step (\ref{z-step}) in Algorithm \ref{alg:mragg} we end up with a $\hz^+$ which is consistent with $\rhob_0$ but is partially ordered. When such a partially ordered $\hz^+$ is used as an input for estimating $\hr^+$ at step (\ref{r-step}) of Algorithm \ref{alg:mragg}, the final output for $\hr^+$ may have a total ordering $\rhob_1$ which is consistent with the partial ordering specified by $\hz^+$, but not necessarily consistent with the total ordering $\rhob_0$. Therefore, between time $t$ and time $t+1$, the algorithm ends up moving from the ordering $\rhob_0$ to the ordering $\rhob_1$.\\

%Suppose at time $t= t_0$ we start with some initial total ordering for $\hr$ and $\hz$ and after step \ref{z-step} of algorithm \ref{alg:mragg}, we end up with a partially ordered $\hz^+$. On using this partially ordered $\hz^+$ as the rank order input to MR in step \ref{r-step} of Algorithm \ref{alg:mragg}, an intermediate step may involve a permutation estimation that imposes a total ordering on the estimated $\hr^+$ that is consistent with the partial ordering on $\hz^+$, but not necessarily with the initial total ordering on $\hz$. Thus, although we may start with some ordering at $t=t_0$, we can end up with a very different ordering at step $t=t_0 + 1$. 
Subsequently, we shall show with experiments on synthetic data that this is indeed the case. Starting from an initialisation that does not reflect the true ordering, the $\hr$ and $\hz$ obtained by our algorithm are allowed to move between permutations till they converge to a stationary point corresponding to the exact ordering.

\subsection{Extensions}\label{sec:extensions}

While the methods outlined in this manuscript used GLM's and Bregman Divergences, our formulation is much more general. Specifically, techniques like MR were used to illustrate one of many possible ways of jointly modeling rank scores with object features, and other methods from the supervised LETOR or rank aggregation literature can be easily incorporated into our framework by changing the cost function and optimisation algorithm. Similarly, our methods can be extended to many other contexts, including partial orderings that can be either learned at each step iteratively \cite{MR}, or by using isotonic regression algorithms \cite{iso} when specified using directed acyclic graphs \cite{dag}, or using additional constraints when specified using implicit feedback \cite{rendle}. Finally, adding supervision to the framework is straightforward -- simply add strict wide-margin (linear half-space) constraints to the margin equipped formulation described in section \ref{sec:margin}.

%Spurious rank orders
%Extensions
%Choice of Phi
\begin{comment}
\begin{figure*}[!htb]
\centering
\subfloat[subfig1][Kendall-Tau distance versus Iteration for Gaussian]{
\includegraphics[width=0.5\textwidth, height = 6.0cm]{SimKDT_Gauss0}
\label{fig:kdtgauss}}
\subfloat[subfig2][Kendall-Tau distance versus Iteration for Poisson]{
\includegraphics[width=0.5\textwidth, height = 6.0cm]{SimKDT_Poiss0}
\label{fig:kdtpoiss}}
\caption{Synthetic Data: Kendall's tau metric (higher is better) against iteration for (a) Gaussian (b) Poisson model. Our methods (MR w/ Rank Score and MR w/ Features) exactly recover the true rank ordering within a few iterations}
\label{fig:kdtSim}
\vspace*{-0.5cm}
\end{figure*}
\end{comment}

\begin{figure*}%[!htb]
\centering

\begin{tabular}{cc}

\hspace*{-0.7cm} \subfloat[subfig1][Kendall-Tau distance versus Iteration for Gaussian]{
\includegraphics[width=0.5\textwidth, height = 6.0cm]{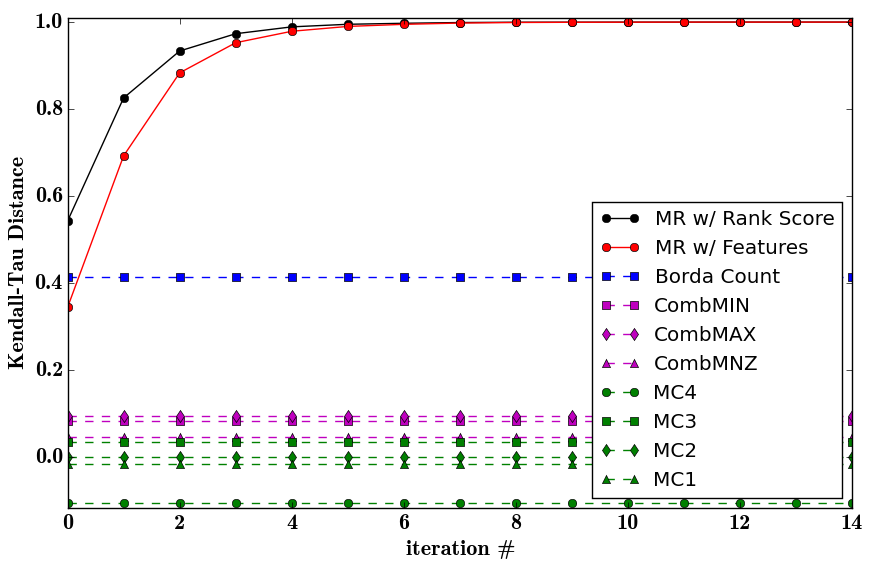}
\label{fig:kdtgauss}} &
\hspace*{-0.2cm} \subfloat[subfig2][Kendall-Tau distance versus Iteration for Poisson]{
\includegraphics[width=0.5\textwidth, height = 6.0cm]{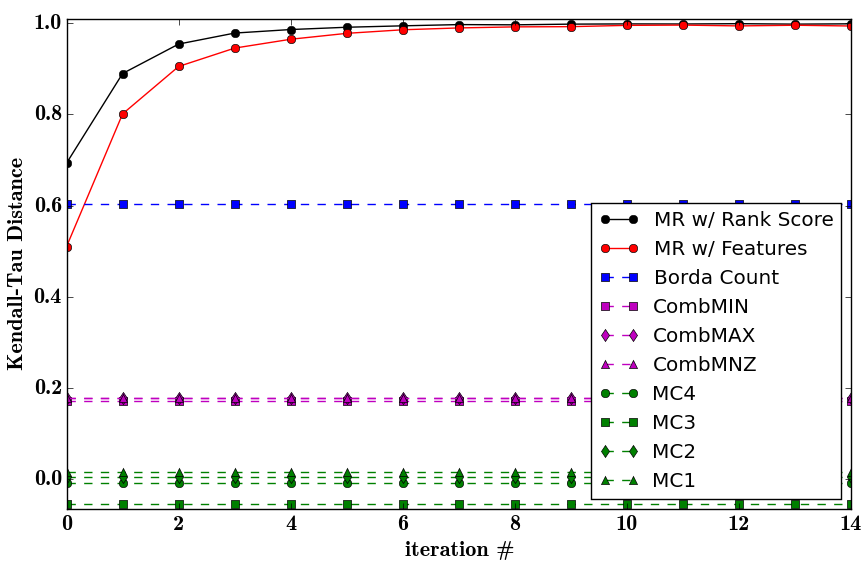}
\label{fig:kdtpoiss}} \vspace*{0.5cm} \\

\hspace*{-0.7cm} \subfloat[subfig3][Spearman's rho against iteration for Gaussian model]{
\includegraphics[width=0.5\textwidth, height = 6.0cm]{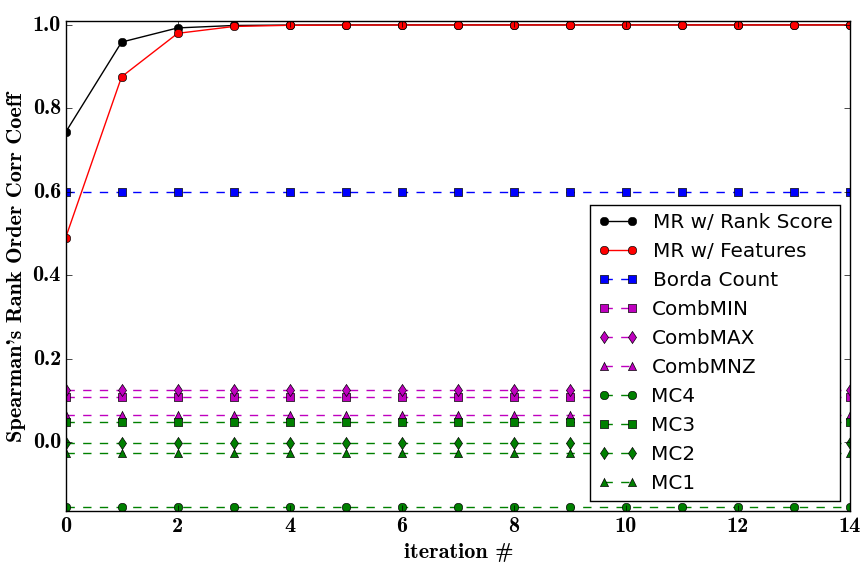}
\label{fig:sprgauss}} &
\hspace*{-0.2cm} \subfloat[subfig4][Spearman's rho against iteration for Poisson model]{
\includegraphics[width=0.5\textwidth, height = 6.0cm]{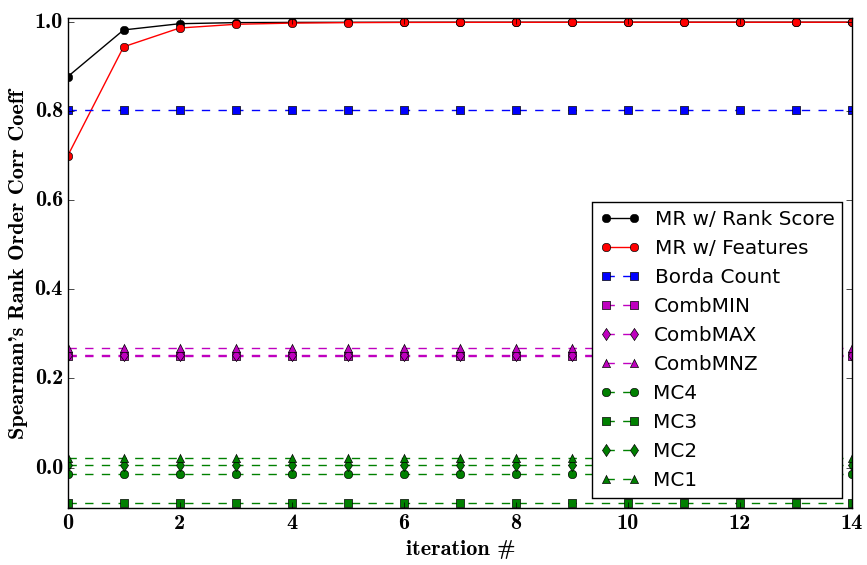}
\label{fig:sprpoiss}} \vspace*{0.5cm} \\

\hspace*{-0.7cm} \subfloat[subfig5][NDCG@K versus K for Gaussian]{
\includegraphics[width=0.5\textwidth, height = 6.0cm]{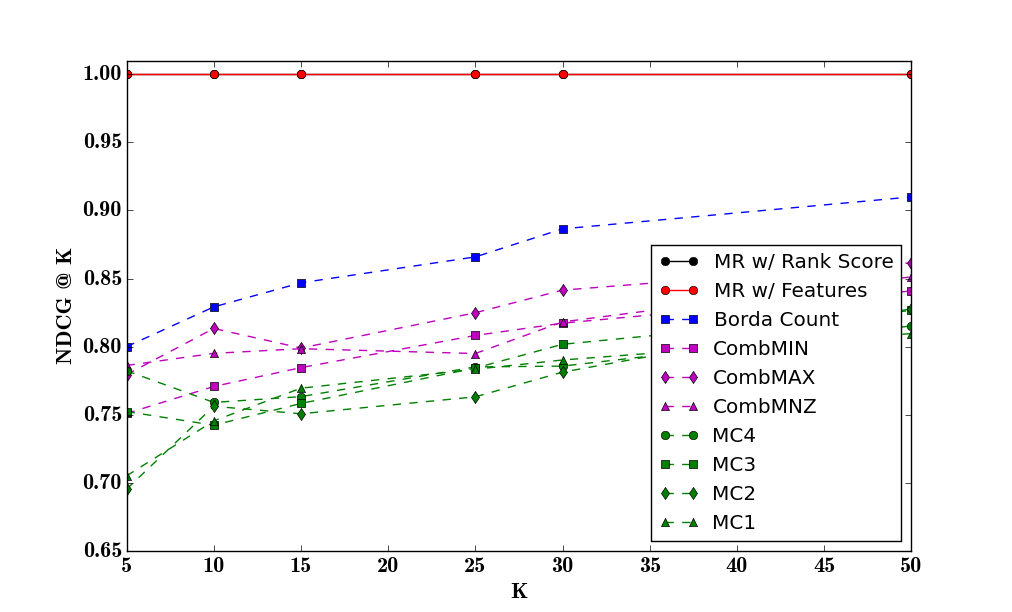}
\label{fig:ndcggauss}} &
\hspace*{-0.2cm} \subfloat[subfig6][NDCG@K versus K for Poisson]{
\includegraphics[width=0.5\textwidth, height = 6.0cm]{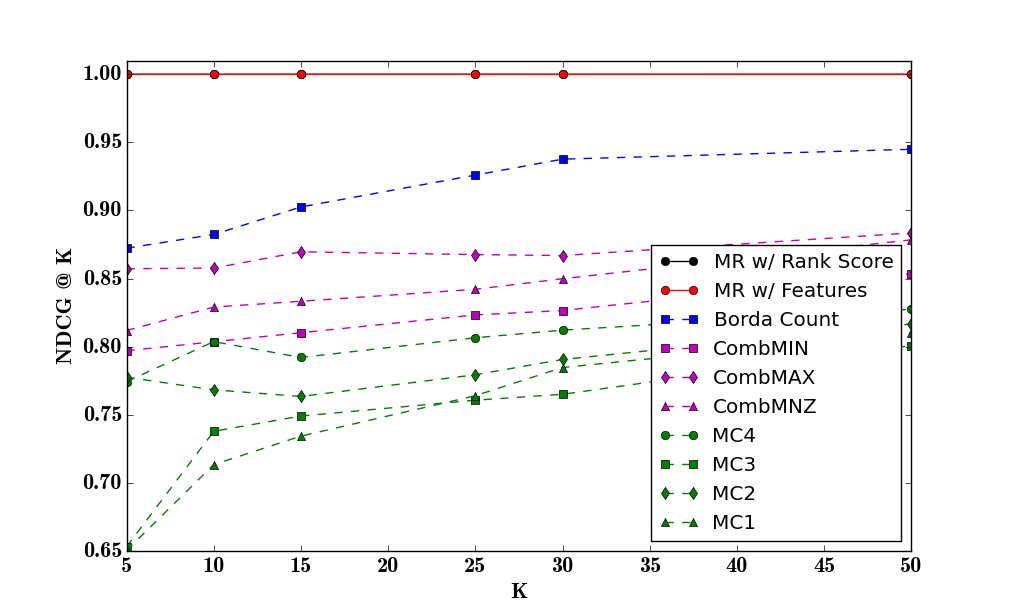}
\label{fig:ndcgpoiss}}  \vspace*{0.5cm} 

\end{tabular}

\caption{Synthetic Data: Kendall-Tau, Spearman's Rho and NDCG@K vs K against iterations of the algorithm for Gaussian [figures (\ref{fig:kdtgauss}), (\ref{fig:sprgauss}), (\ref{fig:ndcggauss}) respectively] and Poisson models [figures (\ref{fig:kdtgauss}), (\ref{fig:sprpoiss}), (\ref{fig:ndcgpoiss}) respectively] \\ Results show that our method can exactly recover the true rank order even from corrupted rank lists within only a few iterations. In contrast none of the baseline rank aggregation methods can recover the true ordering.}

\end{figure*}

\begin{figure*}%[!htb]
\centering

\begin{tabular}{cc}

\hspace*{-0.7cm} \subfloat[subfig1][NDCG@K versus K on the OHSUMED dataset]{
\includegraphics[width=0.5\textwidth, height = 6.0cm]{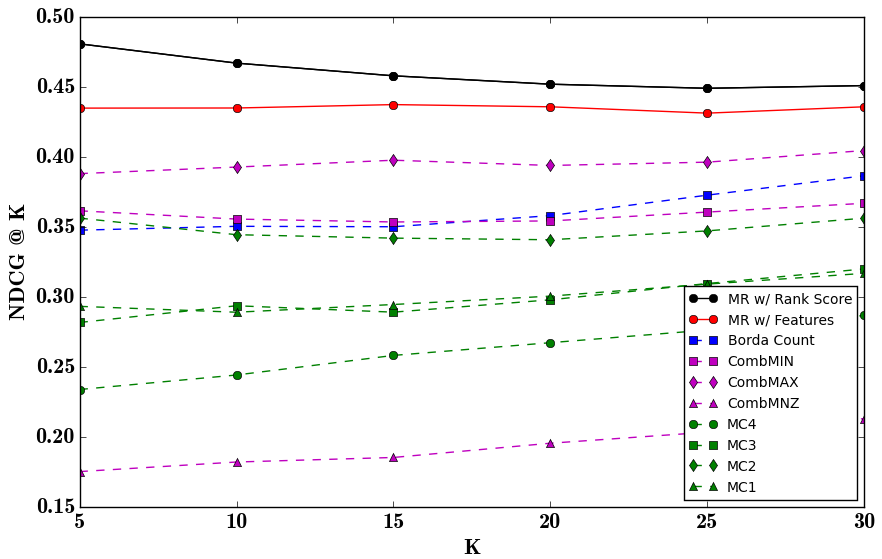}
\label{fig:ohsumed_1}} &
\hspace*{-0.2cm} \subfloat[subfig2][NDCG@K vs K on OHSUMED with augmented rank list]{
\includegraphics[width=0.5\textwidth, height = 6.0cm]{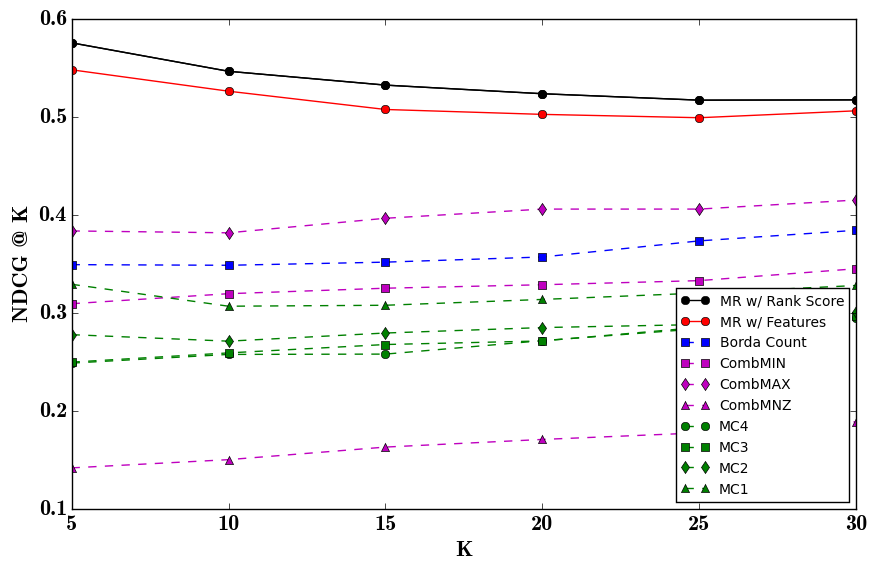}
\label{fig:ohsumed_aug}} \vspace*{0.5cm} \\

\hspace*{-0.7cm} \subfloat[subfig3][NDCG@K versus K on the MQ2008 dataset]{
\includegraphics[width=0.5\textwidth, height = 6.0cm]{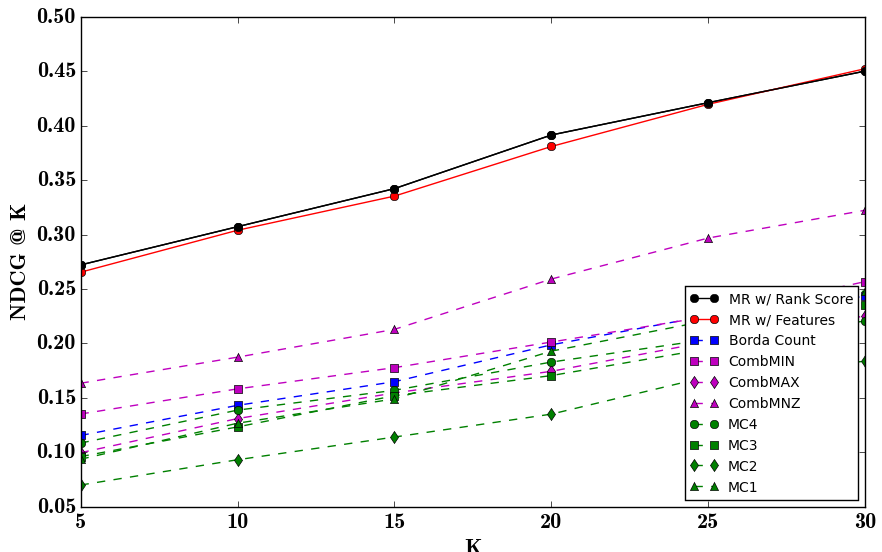}
\label{fig:mq2008_1}} &
\hspace*{-0.2cm} \subfloat[subfig4][NDCG@K versus K on MQ2008 with augmented rank list]{
\includegraphics[width=0.5\textwidth, height = 6.0cm]{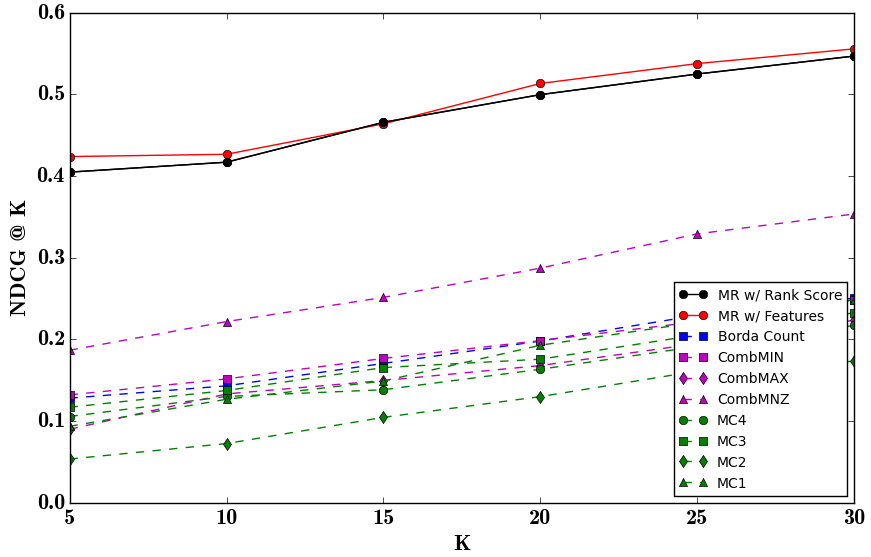}
\label{fig:mq2008_aug}} \vspace*{0.5cm} \\

\hspace*{-0.7cm} \subfloat[subfig5][NDCG@K versus K on the MQ2007 dataset]{
\includegraphics[width=0.5\textwidth, height = 6.0cm]{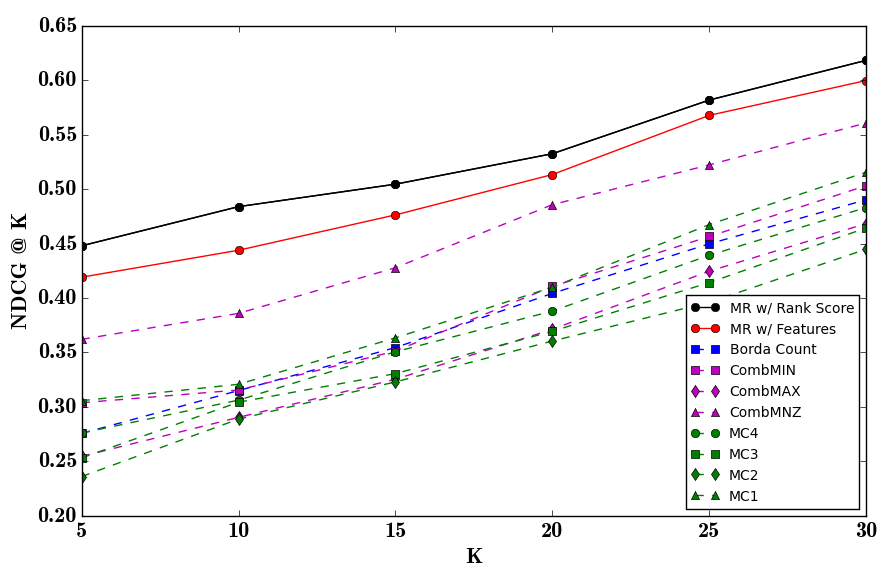}
\label{fig:mq2007_1}} &
\hspace*{-0.2cm} \subfloat[subfig6][NDCG@K versus K on MQ2007 with augmented rank list]{
\includegraphics[width=0.5\textwidth, height = 6.0cm]{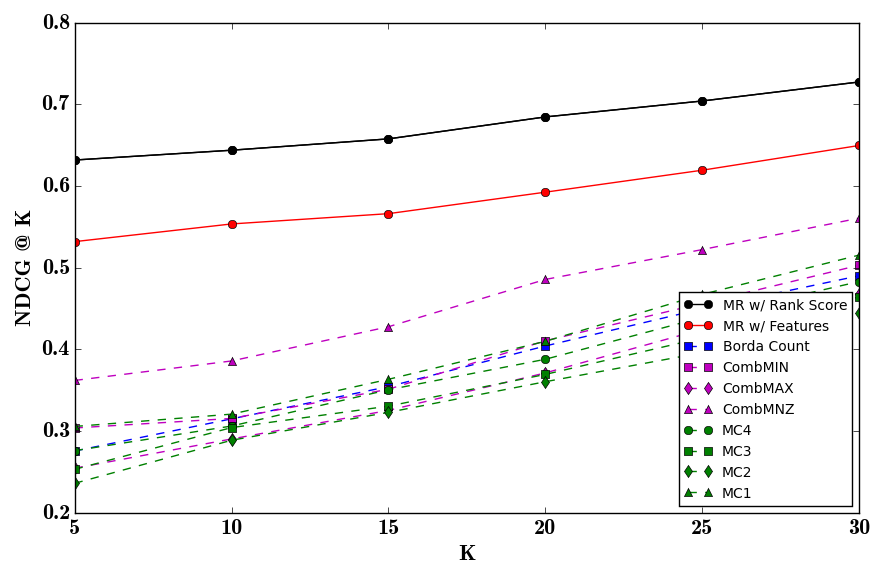}
\label{fig:mq2007_aug}}  \vspace*{0.5cm} 

\end{tabular}
\caption{Real Datasets: NDCG@K versus K (higher is better) averaged across queries OHSUMED, MQ2008, MQ2007 datasets [figures: (\ref{fig:ohsumed_1}), (\ref{fig:mq2008_1}), (\ref{fig:mq2007_1}) respectively], and NDCG@K vs K averaged across queries on the same datasets OHSUMED, MQ2008, MQ2007 with augmented rank lists [figures: (\ref{fig:ohsumed_aug}), (\ref{fig:mq2008_aug}), (\ref{fig:mq2007_aug}) respectively] \\ Results show that our methods outperform standard rank aggregation techniques in general. If the rank lists have "good quality" experts, the improvement in performance is substantial for our method as opposed to baselines}

\label{fig:realdata}
\end{figure*}

\section{Experiments}

We evaluate the performance of our algorithm on both synthetic and real datasets. The metrics used for evaluation are Kendall's tau coefficient \cite{kdt} and Spearman's rho or rank correlation coefficient \cite{spr}, as well as the popular NDCG metric \cite{ndcg} for list-wise ranking. In all three metrics, a higher value indicates better recovery of rank order, and a value of 1.0 indicates exact order recovery. \\

As baselines, we use the classical Borda Count. Among linear combination methods, following \cite{comb2} and \cite{comb3} we use CombMNZ, and also show comparisons to non-linear aggregation methods like CombMIN and CombMAX \cite{comb1}. Further, we also compare against all four Markov Chain based methods presented in \cite{dwork}.

\subsection{Synthetic Data}

We first evaluate on synthetic data where the GLM assumption holds (modulo noise). The rank aggregation framework is evaluated over $n=200$ items for each query\footnote{usually in most real life applications the set of items range between 50 and 300, eg. see the standard datasets in Microsoft LETOR 4.0 \cite{letor4}}. Matrix of feature vectors $\bX$ is generated using the standard multivariate normal distribution and a normally distributed $\omegab^*$ is used to compute the true rank scores ${\boldsymbol \rho}^*$ from $\bX\omegab^*$, according to the generalised linear models corresponding to Gaussian and Poisson distributions respectively, with the appropriate $\bphi$ function (see Appendix \ref{sec:Bregman}). \\%The matrix of expert rank lists $\bR$ as created by concatenating 10 ranked lists consisting of both perturbed versions of the true rank preferences as well as spurious rank lists with no relation to the true rank scores. 

The rank lists consist of vectors generated by randomly corrupting the true rank score ${\boldsymbol \rho}^*$ with different perturbations including translation, and additive or multiplicative noise. Multiple spurious expert rank lists are also generated as vectors constructed from pure random noise. In all, 10 rank lists are concatenated to form $\bR$.\\

$\bX$ and $\bR$ are then used as input to our Algorithm \ref{alg:mragg} with the appropriate Bregman divergence and $\bphi(\cdot)$, and the outputs $\hz$ and $\hr$ are then evaluated\footnote{If the final output is partially ordered, evaluation is done on the total ordering most consistent with the corresponding covariates $\bX\omegab$ for $\hz$ and $\bR\bbeta$ for $\hr$} against ${\boldsymbol \rho}^*$.\\

Figures (\ref{fig:kdtgauss}) and (\ref{fig:sprgauss}) respectively show the value of Kendall's tau and Spearman's rho between the estimated $\hr, \hz$ and the true ${\boldsymbol \rho}^*$ at each iteration of the algorithm, for the setup that uses a Gaussian model. Figures (\ref{fig:kdtpoiss}) and (\ref{fig:sprpoiss}) show the same for a setup generated using a Poisson model. The plots show that in both models, the algorithm is able to exactly recover the original rank ordering (Kendall-tau and Spearman's rho both evaluate to 1.0) within a small number of iterations. In contrast, none of the standard rank aggregation methods are able to recover the true ordering. \\

Figures (\ref{fig:ndcggauss}) and (\ref{fig:ndcgpoiss}) show NDCG@K vs K for the final ordering output by our method, as compared to baseline methods, in a Gaussian setup and a Poisson setup respectively. As opposed to the rank orderings output by our method which always perfectly extracts the top K items for each K, none of the baselines are able to correctly identify the top K relevant items for any value of K.\\
 
%The supplementary material contains figures that corroborates this assertion with corresponding plots for Spearman's rho which also evaluates to 1.0 within a short number of iterations. Furthermore, the supplement contains plots of the NDCG@K versus $K$ evaluated for our output ordering (the value is 1.0 for each $K$) against the performance of the baselines.

\subsection{Real Datasets: MQ2008, MQ2007,\\[5pt] and OHSUMED}

We evaluate our techniques as applied to the more challenging case of real datasets where the GLM assumption may not always hold. We use three complex real world datasets widely used for ranking applications- MQ2008 and MQ2007 from Microsoft's LETOR 4.0 repository \cite{letor4}, and OHSUMED from the LETOR 3.0 repository \cite{letor3}. \\

The MQ2008 and the MQ 2007 datasets from the LETOR 4.0 repository are query sets from the Million Query track of TREC 2008. The LETOR 4.0 repository contains, for each query and associated document set, 46 object features for ranking as well as a set of 25 rank lists. We use the rank lists as our $\bR$ and object features as our $\bX$ matrix respectively. %, with the correspondence matched by query id and document id. 
PageRank and relevance scores computed from different IR methods \cite{bm25,lmir} are also added to the rank lists (they are not used as object features). \\

The OHSUMED dataset is a subset of the MEDLINE database of medical publications, and the standard application involves extraction of relevant documents given a set of medical queries. The rank list matrix is constructed from 15 columns of the dataset that contain relevance scores computed using the BM25 score\cite{bm25} and different IR methods based on language models \cite{lmir}. The remaining 30 columns are used as object features. \\

In each of these datasets, we compare against ground truth which is available for each query and associated item as a relevance score that goes from 0 (not relevant) to 2 (most relevant) (ground truth is not used for learning). Additional details about the datasets including feature lists are available in \cite{letor4} and \cite{letor3} respectively. Note that for each experiment, object attributes and rank lists use disjoint sets of columns.\\

We compare the performance of different methods using the NDCG@K metric applied to this relevance score, averaged across queries for each K. The results are shown in figures (\ref{fig:ohsumed_1}), (\ref{fig:mq2008_1}), (\ref{fig:mq2007_1}) for OHSUMED, MQ2008, MQ2007 respectively. The plots show that for all three datasets, even though the GLM assumption may not necessarily hold, our method nevertheless outperforms standard rank aggregation methods which do not take into account object features.\\

%Interestingly, the performance of our method improves substantially if one or more of the rank lists are accurate% (for example, obtained using a ranking function trained with ground truth supervision)
%, while standard rank aggregation methods only show marginal change in performance on the same. We show the corresponding plots in the supplement where the rank matrix is augmented with orderings output by a rank function trained on ground truth using the methods in \cite{MR}.

An interesting phenomenon that was observed with real datasets was that when at least one of the rank lists are of ``high quality" in the sense that it is learned with supervision in the form of true relevance scores, the performance of our algorithm improves substantially. This effect is not seen on the baselines which only show marginal improvement compared to their performance on set of rank lists which has not been augmented with high quality lists.\\

We performed a similar experiments as above, except we augmented our rank lists from experts with the relevance scores output by a ranking function learned from training data using the methods in \cite{MR}. The NDCG@K versus K plots for augmented rank lists are shown for the OHSUMED dataset in fig. (\ref{fig:ohsumed_aug}), for the MQ2008 dataset in fig. (\ref{fig:mq2008_aug}),  and for the MQ2007 dataset in fig. (\ref{fig:mq2007_aug}).\\ %The original plots are included for side by side comparison. 

Side by side comparison with original plots show that when the rank lists contain at least one list of high quality, our algorithm returns an aggregated ordering in which the NDCG score shows a marked improvement, while standard methods fail to exploit the augmented rank lists to any substantial degree. Note that information about which rank lists are of high quality is not available to any of the algorithms in these experiments.\\ %Both our method and the baselines use the augmented rank lists in the same way as they would a rank list without the high quality components.

These experimental results suggest that if the set of rank lists are generated by sources whose expertise levels lie on a wide spectrum, our framework is nevertheless able to use more information from rank lists with higher credibility, and recover an ordering which matches the ``true" ranking to a greater degree as compared to rank aggregation methods which depend solely on the rank lists and are blind to object features.

\section{Conclusion}

In this manuscript we introduced a novel rank aggregation scheme that augments expert rank lists with information from object features to bypass the various issues that plague the standard rank aggregation setup, and in the process obtain more accurate and robust aggregated rank scores. Experiments on synthetic data and on real datasets indicate that using object features can result in significant improvement in performance, more so when the rank lists contain orderings generated by sources with genuine expertise. Future work would cover theoretical analyses of our scheme, including statistical guarantees and extensions to more general models for ranking and rank aggregation.

\bibliographystyle{abbrv}
\bibliography{rank}
\appendix

\section*{APPENDIX}

\section{Bregman Divergences}\label{sec:Bregman}
The matching loss functions associated with learning GLM parameters are distance-like functions called Bregman divergences, which are generalisations of square loss. Let $\bphi : \Theta \mapsto \mathbb{R}$ be a strictly convex, closed function on a convex domain $\Theta \subseteq \mathbb{R}^m$. Suppose $\bphi$ is differentiable on int($\Theta$). Then, for any $\bx, \by \in \Theta$, the Bregman divergence $D_\bphi(\cdot \| \cdot)$ between $\by$ and $\bx$ corresponding to the function $\bphi$ is defined as
\begin{equation*}
D_\bphi(\mathbf{y} \| \mathbf{x}) \triangleq \bphi(\mathbf{y}) - \bphi(\mathbf{x}) -  \langle \nabla\bphi(\mathbf{x}), \mathbf{y - x} \rangle
\end{equation*}
Bregman divergences are convex in their first argument. Although strictly speaking they are not a distance metric, they satisfy many properties of metrics, for example $D_\bphi(\mathbf{y} \| \mathbf{x}) \geq 0$ and $D_\bphi(\mathbf{y} \| \mathbf{x}) = 0$ if and only if $\mathbf{y = x}$. %MR uses Bregman divergences defined on convex functions that separate out into sums of identical scalar convex functions over each component of the vector. 
Many standard distance-like functions like Square loss and KL-divergence are members of this family (see Table \ref{tb:breg}).\\

There is a one-one correspondence between each GLM and each Bregman divergence via the convex function $\bphi(\cdot)$, and parameter estimation in a GLM with given object features $\bX$ and target variable $\hz$ is equivalent to finding the minimiser for $D_\bphi\left(\mathbf{z} \| \gpi(\mathbf{X \omegab})\right)$ over $\omegab$, where $\bphi(\cdot)$ is the convex function associated with the particular GLM used (refer to \cite{banerjee} for a detailed exposition on the relationship between Bregman Divergences and GLM's). %\vspace*{-0.2cm}

\setlength{\intextsep}{10pt} 
%\begin{comment}
\begin{table}[h]
\begin{center}
\begin{tabular}{| c | c |}
\hline
{$\bphi(\mathbf{x}$)}  &{$D_\bphi(\mathbf{y \| x})$} \\
\hline 
$\frac{1}{2}\|\mathbf{x}\|^2 $  & $\frac{1}{2}\|\mathbf{y - x}\|^2$ \\ \hline
\specialcell{$\sum_i (x^{(i)} \log x^{(i)})$ \\ $\mathbf{x} \in$ Prob. Simplex } & \specialcell{KL($\mathbf{y \| x}$) = \\ $\sum_i \left(y^{(i)} \log(\frac{y^{(i)}}{x^{(i)}})\right)$} \\ \hline
\specialcell{$\sum_i \left(x^{(i)} \log x^{(i)} - x^{(i)}\right)$ \\ $\mathbf{x} \in \mathbb{R}^n_+$}  & \specialcell{GI($\mathbf{y \| x}$) = \\ $\sum_i y^{(i)} \log (\frac{y^{(i)}}{x^{(i)}}) - y^{(i)} + x^{(i)}$}\\ \hline
\end{tabular}
\caption{\footnotesize Examples of Bregman Divergences} \label{tb:breg}
\end{center}
\end{table}
%\end{comment}

\section{Monotone Retargeting}\label{sec:MR}

%Learning to rank or LETOR is a supervised learning problem for a ranking model in which the relevance score or rank score for an object is a function of the object features. 

Monotone retargeting \cite{MR, memr} or MR is a LETOR framework that models ranking functions of object features on monotonic transformations of given relevance scores, and learns both the ranking function as well as the best monotonic transformation.
%Monotone retargeting or MR \cite{MR, memr} is a framework for LETOR that assumes that the true rank scores are isotonic with a monotonically transformed linear function of the object features. 
The paper \cite{MR} solves the LETOR problem using MR by modeling the monotonically transformed rank score $\hz$ as a generalised linear model over object features $\bX$ and learns the GLM parameter $\omegab$ that best fits the data over all possible monotonic transformations $\hz$ of the given ``true" relevance scores ${\boldsymbol \rho}^*$.\\

The LETOR setup used by MR is the following. Consider a single query. Suppose for a given set of $n$ items $\{V_1, V_2, \cdots V_n\}$ to be ranked, $\bX \in \bbR^{n\times p}$ is the matrix of object features. Suppose the supervision is provided as a rank score vector ${\boldsymbol \rho}^* \in \bbR^n$. The objective of MR is to find a monotonic transformation of the given rank scores $\hz \sim_\downarrow {\boldsymbol \rho}^*$ and a real valued parameter $\omegab \in \bbR^p$ which minimises the Bregman divergence $\dppi{\hz}{\bX \omegab}$. The optimisation problem, therefore, is the following
\begin{equation} \label{eq:mr}
\begin{aligned}
&\underset{\hz, \omegab}{\text{min}} & & D_\bphi\left(\mathbf{z} \| \gpi(\mathbf{X \omegab})\right) \\
& \text{s.t. } & & \hz \sim_\downarrow {\boldsymbol \rho}^*
\end{aligned}
\end{equation}

Given ${\boldsymbol \rho}^*$ and $\bX$, this problem is separately convex in $\hz$ and $\omegab$, and with appropriate regularisation can also be made jointly convex in the two variables\cite{memr}. A top level description of the algorithm as adapted from \cite{MR} is presented as Algorithm \ref{alg:mr} in the main manuscript. In particular, the steps involved include a standard GLM parameter estimation for $\omegab$ and the pool adjacent violators algorithm \cite{pav}, widely used in isotonic regression, for $\hz$. If ${\boldsymbol \rho}^*$ is partially ordered, an intermediate step involves the estimation of a total ordering consistent with the partial ordering specified by ${\boldsymbol \rho}^*$. Each step of the algorithm has been widely studied in the literature and off-the-shelf solvers can be used to efficiently iterate through each step of the algorithm to converge to a stationary point. See \cite{MR} for a more detailed discussion on the properties of the solution obtained via this framework, and efficient algorithms for optimisation.\vspace*{-0.2cm}% (see \cite{MR} and \cite{memr} for more details). 

%\begin{algorithm}
%\caption{LETOR with Monotone Retargeting}\label{alg:mr}
%\begin{algorithmic}[1]
%\Procedure{MR}{$\bphi, \bX, {\boldsymbol \rho}$}%\Comment{The g.c.d. of a and b}
%\State Initialise $\omegab, \hz$
%\While {not converged}
%\State Solve for $\hz^+$ using PAV $$\hz^+ = \text{arg}\min_{\hz \sim_\downarrow {\boldsymbol \rho}} \dpi{\hz}{\bX \omegab}$$
%\State Solve for $\omegab^+$ a std GLM param estimation
%$$\omegab^+ = \text{arg}\min_{\omegab} \dpi{\hz^+}{\bX \omegab}$$
%\State Update variables $(\hz, \omegab) = (\hz^+, \omegab^+)$
%\EndWhile
%%\State 
%\State \textbf{return} $\hz, \omegab$%\Comment{Sparse map is }
%\EndProcedure
%\end{algorithmic}
%\end{algorithm}

\section{Margin Equipped Monotone Retargeting}\label{sec:memr}

A margin equipped variation of this problem, MEMR\cite{memr}, avoids degenerate stationary points by enforcing a margin between the relevance score between any two items in the setup. That is, suppose ${\boldsymbol \epsilon} = [\epsilon_1, \epsilon_2, \cdots \epsilon_{n-1}] \in \bbR^{n-1}_+$, where each $\epsilon_j \geq 0$. Suppose the specified rank score vector ${\boldsymbol \rho}^*$ is ordered as $(\tau_1, \tau_2, \cdots \tau_n )$, where each $\tau_j \in \{1, 2, \cdots, n\}$. That is, ${\boldsymbol \rho}^{\tau_j} \geq {\boldsymbol \rho}^{\tau_{j+1}}$ for each $j = 1, 2, \cdots (n-1)$. %Without loss of generality, assume that the rank score vector is in increasing order from top to bottom $\hr \sim_\downarrow \bbR^n_\downarrow$. 
Then the reformulated optimisation problem used in MEMR is as follows
\begin{equation} \label{eq:memr}
\begin{aligned}
&\underset{\hz, \bbeta}{\text{min}} & & D_\bphi\left(\mathbf{z} \| \gpi(\mathbf{X \bbeta})\right) \\
& \text{s.t. } & & \hz \sim_\downarrow {\boldsymbol \rho}^*\\
&  & & \hz^{\tau_{j}} - \hz^{\tau_{j+1}} \geq \epsilon_j\ \ \forall\ j = 1, 2, \cdots n-1
\end{aligned}
\end{equation}

Since the additional constraints are convex half-space constraints over $\hz$, most of the standard properties of monotone retargeting are maintained (see \cite{memr} for a more detailed analysis as well as optimisation algorithms).

\end{document}

%% file: econfmacros.tex
\def\beq{\begin{equation}}
\def\eeq#1{\label{#1}\end{equation}}
\def\eeqn{\end{equation}}

\def\beqa{\begin{eqnarray}}
\def\eeqa#1{\label{#1}\end{eqnarray}}
\def\eeqan{\end{eqnarray}}

\let\bar=\overbar

\def\Dslash{\not{\hbox{\kern-4pt $D$}}}
\def\dslash{\not{\hbox{\kern-2pt $\del$}}}

\def\msb{{\bar{\ssstyle M \kern -1pt S}}}